# Evaluation of a Foundational Model and Stochastic Models for Forecasting Sporadic or Spiky Production Outages of High-Performance Machine Learning Services

Keun Soo Yim

*Abstract*—Time series forecasting models have diverse real world applications (e.g., from electricity metrics to software workload). Latest foundational models trained for time series forecasting show strengths (e.g., for long sequences and in zero-shot settings). However, foundational model was not yet used for forecasting rare, spiky events, i.e., a challenging target because those are a corner case of extreme events. In this paper, we optimize a state-of-the-art foundational model to forecast sporadic or spiky production outages of high-performance machine learning services powering billions of client devices. We evaluate the forecasting errors of the foundational model compared with classical stochastic forecasting models (e.g., moving average and autoregressive). The analysis helps us understand how each of the evaluated models performs for the sporadic or spiky events. For example, it identifies the key patterns in the target data that are well tracked by the foundational model vs. each of the stochastic models. We use the models with optimal parameters to estimate a year-long outage statistics of a particular root cause with less than 6% value errors.

## I. INTRODUCTION

Various kinds of applications we use in our daily lives utilize high-performance machine learning services. These services, for example, are used to analyze and process various types of user-created contents (UCC), such as text, images, and videos, often leveraging cloud computing for scalable operations. As billions of users rely on such services, production outages can significantly impact a large number of users around the globe. Specifically, a user-visible outage of a cloud-based machine learning platform can prevent users from accessing personalized contents, using intelligent features, creating personalized content, and benefiting from real-time analytics. This can result in the brand value damages, advertisement revenue losses, and reputation damage for the machine learning service providers. Moreover, some users can complain on the social media, show desperation, and look for alternative services [1][2]. Hence, mitigating user-visible production outages in these high-performance machine learning services is a critical technical challenge in the era of generative artificial intelligence (GenAI) and UCC.

The initial steps in controlling the production outages of the high-performance machine learning services are to: measure the historical outage statistics and estimate the future statistics until the end of a specific time period (e.g., project period or fiscal year). One may adopt an existing software reliability growth model (SRGM) [3][4][5][6] that is originally devised for standalone computers where it is assumed the total defect count is fixed. However, our curve fitting analysis in this paper shows one-parameter models do not accurately estimate the outages of a software service developed and released by using a highly iterative development method (e.g., agile [7]). That is in part because defects are continuously introduced and removed in agile software where the defect birth and death rates are not identical. That is, there are more than one parameters deciding the defect counts in agile software. Considering the fact that outages can also be caused by operational issues, it suggests us that sophisticated time series forecasting models are necessary to accurately estimate the outages of agile software services.

In this paper, we use a state-of-the-art foundational model (TimesFM [8]) trained on time series data to forecast the outage counts of planet-scale machine learning services powering billions of desktop, mobile, embedded, and IoT client devices. We also use the three classical time series forecasting models (e.g., moving average and autoregressive) as reference models. We use a few techniques (e.g., fine-tuning) to optimize the forecasting accuracy of the considered models.

We use the seven years of the outage statistics of our target services. The first year data is used for training and validation, while the remaining six years of data are used for testing. The initial result shows that the foundational model has the highest accuracy (e.g., 1.3-12.4% higher than the second best model) when it is used to forecast the monthly total outage counts (i.e., short-term, single-step forecasting).

We breakdown the outages into sub-categories by using the root cause types. In total, eight root cause types are identified and then used to label the outages. Each root cause type thus has time series data for its monthly outages. We evaluate the forecasting accuracies of the four forecasting models. For each root cause type, the most accuracy model varies especially because the target events are sporadic or spiky (e.g., with a

K. S. Yim is with Google, Alphabet Inc., attn. yim, 1600 Amphitheatre Pkwy Mountain View, CA 94043, USA (e-mail: yim@google.com).

high variance but without any prominent periods).

We analyze the results to better understand how each of the evaluated models performs for the targeted sporadic or spiky events. It identifies the key patterns in our target data that are well or not well tracked by each of the used models with the varying lookback parameter values. Finally, we use the models with optimized parameters to estimate the monthly outages of the seventh year of a specific root cause type (i.e., long-term, iterated multi-step forecasting). The best model always has <6% estimation errors although it is done up to 12 months ahead.

The main contributions of this paper are as follows:
- This is the first work adopting a foundational model to forecast sporadic or spiky software reliability time series data (i.e., production outages), as far as we know.
- We breakdown outages into sub-categories using the root cause type and formulate the software reliability modeling as a time series forecasting problem.
- We show the used foundational model is more accurate than the considered classical models for overall outage counts. We also show neither a model nor a lookback parameter value always shows the highest accuracy for each of the considered 8 root cause types.
- We characterize and discuss when foundational model shows the better forecasting accuracy than the classical stochastic models and vice versa.
- Overall, this work shows the value and potential of a foundational model in software reliability modeling.

The rest of this paper is organized as follows. Section II reviews SRGMs. Section III describes our methodology and characterizes our target dataset. Section IV conduct a curve-fitting study and shows the necessity of sophisticated models. Section V presents the forecasting models and techniques used in this study. Section VI analyzes the evaluation results. Section VII reviews the related work. Section VIII concludes.

## II. BACKGROUND

This section reviews the foundational software reliability models that estimate, $p(t|S)$, the probability of target software ($S$) causing a failure by an operating time $t$. The software reliability is $R(t|S) = 1 - p(t|S)$. This section also discusses the challenges in using those models for agile software, which is updated in a highly iterative manner (e.g., daily or weekly).

Software reliability growth models (SRGMs) are statistical models for predicting the failure rates of a software system, given the historical failures (or found defects) of the system. SRGMs usually make assumptions about the defect discovery and removal processes of the target software. The parameters of both processes are derived from the statistics obtained in the post-development, formal testing phase. In practical, it often uses the statistics from the final testing phase as the historical failures because they are a good indicator of the production reliability (e.g., without the final testing phase). Another typical assumption of SRGMs is that the number of defects (or failures) in a target software system is finite. It thus only models how the software reliability grows (but never decreases) over time as the development and testing progress.

**Target Metrics.** SRGMs are classified into two mutually, non-disjoint sub-classes based on the prediction target metric.

*Mean Time between Failures (MTBF).* This class of SRGMs uses a probability density function, $PDF_i(t)$, to model the time ($t$) between failure $i-1$ and failure $i$. Typically, time is the wall clock time for real projects and the CPU execution time for small projects. The expected value is then $E[t] = \int_0^\infty t \cdot PDF_i(t)dt$. The parameters of $PDF_i(t)$ are estimated using the observed intervals between the previous failures: $t_1, t_2, \ldots, t_{i-1}$. It is typically done by using the maximum likelihood method (to take advantages of the asymptotic normality, asymptotic efficiency, and invariance) or the least squares method. The software reliability is modeled as the probability that the time to the next failure will be more than a certain value ($x$). That is, $R(x) = P(t > x) = \int_x^\infty PDF_i(t)dt$.

*Failures in a Time Interval (Failure Rate).* This class uses another $PDF_i(x)$ where $x$ is the random variable for the failure count. The parameters of $PDF_i(x)$ are estimated based on the failure counts ($f_1, \ldots, f_{i-1}$) in the previous test intervals. The expected value is $E[x] = \int_0^\infty x \cdot PDF_i(x)dx$. Here, $x(t)$ is the number of failures by time $t$ and satisfies $\lim_{t \to \infty} x(t) < \infty$ with the finite failure count assumption.

**Models.** There are more than a hundred of SRGM models [3][4][5][6]. We review the following eight fundamental SRGMs [9][10][11][12] using the parameters: $K$ is the total number of defects (or software faults) initially in the target software system; and $t$ is time between discovery of $(i-1)$-th and $i$-th failures. It shows that many fundamental SRGMs are based on a single-parameter, standard distribution (e.g., exponential or Poisson).

*Exponential Model.* This model is for a defect rate process that declines monotonically to an asymptote. $PDF$ at time $t$ is $K\lambda e^{-\lambda t}$, where $\lambda$ is the defect discovery rate (or hazard rate) per fault. Then, $CDF(t) = K[1 - e^{-\lambda t}]$. It is a special case of the Weibull distribution model with the shape parameter of 1.

Jelinski-Moranda (J-M) model assumes: the fault detection rate is proportional to the current residual faults; all failures are equally likely to occur and are independent of each other; each failure is of the same order of severity as any other failure; the failure rate remains constant over the interval between failure occurrences; during tests, the software operates like it does in the production; and faults are instantly corrected without any new faults introduced. The defect discovery rate is: $Z(t_i) = \Phi(K - (i-1))$ where $\Phi$ is the proportionality constant. Then, the random variable $X_i$ for $t$ is:
$f(X_i) = Z(t)e^{-Z(t)X_i} = \Phi(K - (i-1))e^{-\Phi(K-(i-1))X_i}$
because the assumed failure rate is constant, i.e., exponential distribution.

While the J-M model assumes the perfect debugging, fixing a defect may introduce new defects in practice. Goel-Okumuto imperfect debugging model thus uses the hazard function of $Z(t_i) = [K - p(i-1)]\lambda$ where $p$ is the probability of imperfect debugging.

*Nonhomogeneous Poisson Process (NHPP) Model.* For any time period $(t, t + \Delta t)$, it assumes $\mu(t, t + \Delta t) - \mu(t) = b(v_0 - \mu(t))\Delta t + O(\Delta t)$ where $b$ is the proportionality constant $(= \lambda_0/v_0)$ and $\lim_{\Delta t \to 0} O(\Delta t)/\Delta t = 0$. As $\Delta t \to 0$, the mean function, $\mu(t)$, satisfies: $d\mu(t)/dt = 1 - (\lambda_0/v_0)\mu(t)$. Under $\mu(0) = 0$, the mean function is: $\mu(t) = \int_0^t 1 - (\lambda_0/v_0)\mu(x)dx = v_0(1 - e^{-\lambda_0/v_0 \cdot t})$. The probability that the cumulative number of failures, $K(t)$, is less than $k$ is $p(K(t) \leq k) = \mu(t)^k/k! \cdot e^{-\mu(t)}$. The failure intensity is: $\lambda(\tau) = \lambda_0 e^{-\lambda_0/v_0 \cdot \tau}$. Thus, $\lambda(\mu) = \lambda_0(1 - \mu/v_0)$.

Goel and Okumoto model [9] is based on an NHPP. Its mean value function is: $H(t) = m(t) = a[1 - e^{-bt}]$ for $b > 0$. Here, $b$ is error detection rate per error. $M(t)$ is a constant error detection rate function because $d(t) \equiv d_m(t) = b$ for $t > 0$. The number of remaining software errors in the system at time $t$ is: $\overline{K}(t) = K(\infty) - K(t)$. That is, $E\{\overline{K}(t)\} = a \cdot e^{-bt}$.

Delayed S model models both defect detection and defect isolation processes. Due to failure analysis time, there is a notably delay between the first failure observation and reporting. The cumulative detected defect count is S-shaped curve. It is based on NHPP with a different mean value function: $\mu(t) = K[1 - (1 + \lambda t)e^{-\lambda t}]$.

Inflection S model assumes mutual dependence of detected defects. That is, the more we detect, the more undetected failures become detectable (i.e., faults do not occur independently). It is also based on NHPP with a mean value function: $\mu(t) = K \cdot (1 - e^{-\lambda t})/(1 + i \cdot e^{-\lambda t})$ where $i$ is the inflection factor. Let $\{N(t), t \geq 0\}$ be a counting process representing the cumulative number of software faults by time $t$. By definition, $N(0) = 0$. For any finite collection times, the $n$ random variables $N(t_1), N(t_2) - N(t_1), \ldots, N(t_n) - N(t_{n-1})$ are statistically independent. Let $m(t)$ is the s-expected number of failures by time $t$. $m(t)$ is a bounded, non-decreasing function of $t$: $m(t) = 0$ if $t = 0$ and $m(t) = a$ if $t \to \infty$. Here, $a$ is the s-expected number of errors to be eventually detected. From this, we can derive: $m(t + \Delta t) - m(t) = b\{a - m(t)\}\Delta t + o(\Delta t)$ where $o(\Delta t) \to 0$ as $\Delta t \to 0$, meaning the error rate is proportional to the remaining (undetected) errors. Thus, we get: $m'(t) = ab - bm(t)$. Under the boundary condition, we get $m(t) = a(1 - e^{-bt})$. Here, the assumption is when a software failure occurs, its error is immediately removed and no new errors are introduced.

Musa-Okumuto Logarithmic Poisson execution time model models the number of failures per interval, $K(\tau)$. The model is: $P(K(\tau) = k) = [\mu(\tau)]^k/k! \cdot e^{-\mu(\tau)}$, where $k = 0, 1, 2, \ldots$ and $\mu(\tau) = 1/\theta \cdot \ln(\lambda_0 \theta \tau + 1)$ is the expected number of failures observed by time $\tau$. This model also considers that later fixes have a smaller effect on software reliability than earlier fixes and some functions are executed more frequently than others.

*A. Challenges*

**SRGMs were not originally designed for agile software.** Any SRGMs with the finite failure count assumption model the waterfall or spiral software development methods that are common for standalone embedded systems. In the waterfall or spiral method, when a developed software system is sent for the testing phase, the total number of defects is fixed. On the other hand, software developed by using an agile method is continuously extended and updated over-the-air (OTA), such as daily or weekly. As a result, the total number of defects is changing continuously (e.g., fixing existing defects and introducing new defects). Thus, *the total number of defects is unknown and not a constant value in agile software.*

Another characteristic of agile software is that as software development progresses, the tests and other quality techniques are gradually developed and continuously extended. Thus, *the defect detection rate is not a constant but a variable*, although it is often modeled as a constant in some existing SRGMs. It suggests us to test the effectiveness of SRGMs with such assumptions in modeling defects and other reliability statistics of agile software.

**Production Outage Count Estimation Problem.** ~40% of the high severity production outages of many cloud services were due to software defects [13]. The defects causing the production outages included the error detection and handling bugs (31%), data format related bugs (21%), timing bugs (13%), and constant-setting bugs (7%). It implies *production outage counts are indirect measures of software defect counts and consequently the software reliability*. Since the production outages are also due to release, deployment, and other operational issues [14][15], in this paper, we count or *estimate the production outages as a function of the root cause type* to help estimate the respective software defect counts.

### III. METHODOLOGY

This section describes our experimental methodology. The methodology is: (1) to test the effectiveness of the existing single-parameter models in estimating the production outages, and (2) to evaluate the forecasting accuracy of the major types of time series forecasting models in the short- and long-term, univariate forecasting scenarios.

**Target Services.** Our targets are high-performance machine learning services used by billions of worldwide users across a diversity of client devices and networks. The heterogeneity of these devices and the sheer number of global users with a wide range of network conditions make it incredibly difficult to accurately estimate and tightly control production outages for these services.

**Outages Data Set.** Our outages dataset consists of seven years of the production incident statistics of our target service. The dataset is collected for seven years starting from July. Each incident has manually-classified significance level: negligible, minor, medium, major, and huge. Incidents with medium or higher significance (namely, outages) are used in our analysis.

We calculate and use the time series of monthly outages as a target variable. Let us assume $o_i$ is the outage count of the $i$-th month from when the measurement was started. Then, the target time series variable $O = \{o_i\}_{i=1}^{I}$ and $I = 84$ (=7 years × 12 months).

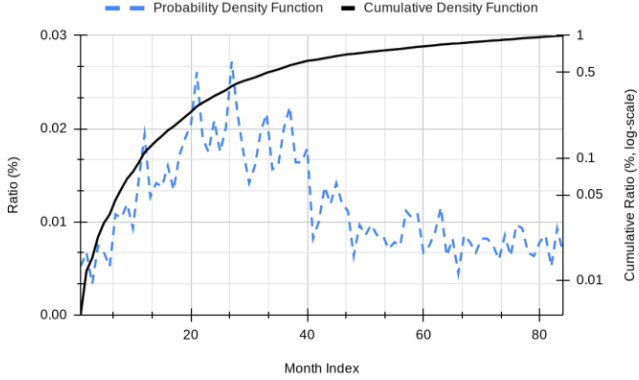

**Fig. 1.** Probability density functions (bar) and cumulative density functions (line) of the monthly production outage ratios (time period: 7 years).

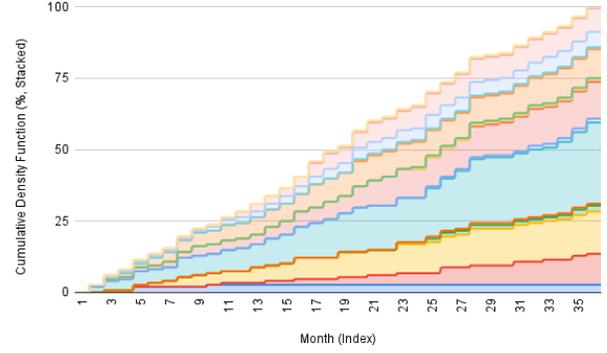

**Fig. 2.** Cumulative density functions (stacked) of the monthly production outage ratios of top-level software components (time period: last 3 years).

**Accuracy Metrics.** The absolute outage counts are relative to various factors, including the target service scale. Thus, we normalize the counts before calculating forecasting accuracy metrics. The normalized outage counts are derived by: $\bar{o}_i = o_i / \sum O$. The forecasted normalized outage counts are notated as: $\hat{O} = \{\hat{o}_i\}_{i=1}^{K}$. As accuracy metrics, we use the normalized mean absolute error ($\overline{MAE}$), normalized mean squared error ($\overline{MSE}$) and root $\overline{MSE}$ ($\overline{RMSE}$) metrics. That is, $\overline{MAE} = (\sum_{i=1}^{I} |\bar{o}_i - \hat{o}_i|)/I$; $\overline{MSE} = (\sum_{i=1}^{I} (\bar{o}_i - \hat{o}_i)^2)/I$; and $RMSE = \sqrt{\overline{MSE}}$. This paper uses the scientific notation for those metric values because the normalized values are small. We also use the mean error percentage, $Error\% = \sum_{i=1}^{I} |\bar{o}_i - \hat{o}_i|/(\bar{o}_i \cdot I)$.

**Breakdown of Outages.** Outages are classified into sub-categories using their root cause types. The eight, considered root cause types are: capacity-, client-, data-, database (DB)-, experiment-, frontend-, machine learning (ML)-, and migration-oriented outages. For every outage, a post-mortem analysis document [14][16] is written by those involved. As part of the analysis, the root cause type(s) are identified.

*A. Characterization*

We characterize the last three years of outages of our target service when the root cause types of outages are classified[1]. ~43% of outages are classified and labeled. The rest does not belong to any of the used eight root cause types. An outage can also have more than one root cause types (i.e., 18% of 43% = ~7.9%), while most of the labeled outages (i.e., 82% of 43% = ~35%) have only one identified root cause type. The capacity root cause type accounts for 4.5% of the total outages; experiment root cause type is for 18.6%; DB type is for 2.3%; frontend type is for 0.8%; data type is for 11.3%; ML type is for 2.9%; client type is for 3.1%; and migration type is for 9.0% in the last 3 years.

**Overall Production Outages.** Figure 1 visualizes the monthly production outage ratios for seven years using the probability density function (PDF) where the cumulative density function (CDF) is shown in log-scale using the right-side vertical axis. It unequivocally demonstrates significant progress in reducing production outages, particularly after 40 months into the data collection period. The PDF graph vividly illustrates the dynamic nature of production outage counts, exhibiting periods of increase, decrease, and relative stability. The presence of these distinct patterns (e.g., upswings, downturns, and both stable high and stable low periods) makes this multi-year dataset an exceptionally valuable and representative target for in-depth analysis and the development of predictive models.

In this preliminary analysis, we only use the last three years of data that is relatively stable and aggregate into sub-periods (e.g., monthly). The averages of the production outage ratios for this three year period are by definition: ~8.3% per quarter, ~2.9% per month, and ~0.6% per week. Their standard deviations decrease as the granularity becomes finer: ~2.3% for the quarterly, ~0.9% for monthly, and ~0.6% for weekly.

The quarterly outage statistics are non-stationary, while the monthly and weekly statistics are stationary. We thus use the *stationary monthly outage counts* in the evaluation. Using an autoregressive model, Augmented Dickey-Fuller (ADF) unit root test determines how strongly given time series can be represented by a non-stationary unit root with time-dependent structures. The p-value of the ADF test is greater than 0.05 for quarterly statistics (~0.101) and is much less than 0.05 for monthly (~2.099×10$^{-6}$) and weekly statistics (~3.891×10$^{-14}$).

In general, the measured *total monthly outages are rare and yet have high variance* (i.e., a type to extreme events [17]), making it difficult to accurately forecast. The visualized CDF in Figure 1 shows that the monthly production outage rate is not a constant value. It also shows a relatively large temporal variations in the monthly outage rate.

**Production Outages for Each Top-Level Component.** We analyze the monthly production outages of each top-level component of the target software service. Figure 2 shows the stacked, cumulative density functions (CDFs). Out of the 15 top-level components, the top 5 components contribute to ~77.0% of the production outages, while the bottom 5 components contribute to ~3.4%. It shows *certain components are noticeably more likely to cause production outages than*

---
[1] For the root cause-specific outage time series, we use the last three years of data to ensure that all root cause types are labeled.

TABLE I
CURVE FITTING RESULTS OF TOTAL OUTAGE COUNTS

| Target Data | Distribution type code | p-value of K-S test | Distribution parameters |
|---|---|---|---|
| Daily CDF | beta | 0.0352 | 0.78859, 0.82402, 1.000 |
| | wrapcauchy | 0.0073667 | 0.09070, 0.0, 0.159155 |
| | uniform | 0.000445 | 0.0, 1.0 |
| | bradford | 0.0003210 | 0.19097, 0.0, 1.0 |
| | mielke | 0.000265288 | 1.00512, 17212847.56942, 1.0 |
| | genpareto | 0.000008574 | -1.04928, 0.0, 1.04928 |
| | foldnorm | 0.000000368 | 1.33898, 0.0, 0.34368 |
| | truncnorm | 0.0000001178 | -0.00019, 0.942350, 1.061177 |
| | nct | 0.00000001421 | 339.968, 1.601769, 0.303427 |
| | powernorm | 0.00000000776 | 0.04285, 0.0, 0.088832 |
| Weekly CDF | beta | 0.92627 | 0.79239, 0.825186, 1.00000 |
| | wrapcauchy | 0.87613 | 0.08855, 0.0, 0.159155 |
| | uniform | 0.62223 | 0.0, 1.0 |
| | mielke | 0.56728 | 1.00720, 127013682.6592, 1.0 |
| | bradford | 0.52067 | 0.13773, 0.0, 1.0 |
| | genpareto | 0.28399 | -1.07145, 0.0, 1.07145 |
| | foldnorm | 0.20764 | 1.3547, 0.0, 0.34263 |
| | nct | 0.11478 | 339.968, 1.61515, 0.302964 |
| | powernorm | 0.11330 | 0.04027, 0.0, 0.086257 |
| | truncnorm | 0.11330 | -3.42e-7, 0.992769, 1.007284 |
| Monthly CDF | wrapcauchy | 0.9999965 | 0.0999999999999992, 0.0, 0.1592 |
| | beta | 0.99168 | 0.83818, 0.749637, 1.000000 |
| | uniform | 0.97999 | 0.0, 1.0 |
| | bradford | 0.97998 | 4.672e-5, 0.0, 1.000000 |
| | mielke | 0.97583 | 1.06751, 61863184.55366, 1.0 |
| | genpareto | 0.96892 | -1.09750, 0.0, 1.097504 |
| | foldnorm | 0.86794 | 1.38458, 0.0, 0.343221 |
| | powernorm | 0.8078454 | 0.0360, 0.0, 0.082528 |
| | truncnorm | 0.7859085 | -0.00011, 0.85892, 1.164247 |
| | nct | 0.7799363 | 339.968, 1.63983, 0.304467 |

TABLE II
CURVE FITTING RESULTS FOR MONTHLY OUTAGES PER ROOT CAUSE TYPE

| Root Cause | Distribution type code | p-value | Derived Parameter Values |
|---|---|---|---|
| Experiment | bradford | 0.9919 | 0.32496, 0, 1.00000000279 |
| | truncnorm | 0.9633 | -0.00008084, 0.9337737, 0, 1.0709 |
| | wrapcauchy | 0.9491 | 0.06092, 0, 0.15915495 |
| ML | wrapcauchy | 0.9494 | 0.10836, 0, 0.15915494 |
| | beta | 0.8564 | 0.83889, 0.859115, 0, 1.000000000 |
| | mielke | 0.7921 | 0.98499, 109677092.8, 0, 1.00000013 |
| Migration | beta | 0.9325 | 0.49325, 0.6501413, 0, 1.00000000 |
| | wrapcauchy | 0.9090 | 0.28123, 0, 0.15915494 |
| | bradford | 0.8530 | 1.18917, 0, 1.0000000002 |

and Poisson.

Curve fitting results are tested by using the Kolmogorov–Smirnov (K-S) test. The p-value of K-S test is summarized in Table I for the best ten distributions along with their derived parameter values. It shows which distribution can capture the target CDFs the most accurately. For all three granularities, the top two distribution types are the same and standout.

**Observation #1.** *The best fitting distributions are beta and wrapped Cauchy. Both distributions have two parameters, suggesting us modeling CDFs of monthly total production outages requires at least two parameters.*

**Curve Fitting Times Series of Each Root Cause Type.** After analyzing the total counts, we now analyze time series of each root cause type. Here, we use the last five years of outage data. We consider the three root cause types: experiment-, ML-, and migration-caused outages. Specifically, we repeat the curve fitting analysis using CDF of monthly outage counts to each of the considered distribution functions. We trim the preceding '0' values and tailing '1' values in the derived CDF.

The p-value of the K-S test is summarized in Table II along with the derived parameter values for the top 3 distribution types. The distribution types with the p-value of higher than the 90% confidence level are: Bradford, truncated normal, and wrapped Cauchy for the experiment type; wrapped Cauchy for the ML type; and beta and wrapped Cauchy for the migration type. Here, Bradford uses 1 parameter, truncated normal uses 4 parameters, and beta and wrapped Cauchy use 2 parameters. While Bradford using 1 parameter shows the highest p-value for one root cause type, only wrapped Caughy using 2 parameters has the high p-values for all three root cause types.

**Observation #2.** *The well-fitted standard distribution (i.e., wrapped Cauchy) for each of the three sub-categories of the production outages uses more than or equal to 2 parameters. Such distributions are effective in part because the root cause-specific outage counts are more extreme events than the total outage counts.*

*the other components.* We found the similar patterns when analyzing the weekly and quarterly statistics.

For the top-level components, the *high variability in the monthly outage ratios per component* can be spotted in Figure 2. The aggregated CDF sub-graphs are not liner and show different paces over time. The variability in the monthly outage ratios per top-level component is more prominent as the analysis granularity becomes finer, e.g., weekly. The standard deviations of the weekly production outage ratios of each of all top-level components are in the range of 0.054% and 0.355%, while the standard deviations are higher than 0.2% for the 5 top-level components.

## IV. ANALYSIS

This section conducts curve fitting studies using standard distributions for the outage statistics of our target software. Our analysis shows estimation models based on one parameter distribution, such as exponential or Poisson, are not good at capturing the outage counts of our target agile software.

**Curve Fitting Using Standard Distributions.** We conduct a curve fitting analysis that tries to fit CDFs of the daily, weekly, or monthly outage counts of our target software to each of the considered standard distribution functions. In total, we consider 56 distribution functions[2] including exponential

---

[2] Alpha, beta, beta prime, Beta-Kappa / Dagum, Bradford, Burr (type III), cosine, double gamma, double Weibull, exponential, exponential power, exponentiated Weibull, fatigue-life (Birnbaum-Saunders), Fisk, folded Cauchy, folded normal, generalized exponential, generalized extreme value, generalized gamma, generalized normal, generalized Pareto, half-logistic, half-normal, hyperbolic secant, inverse Gaussian, inverted gamma, inverted Weibull, Laplace, Levy, log gamma, log-Laplace, logistic, Lomax, Maxwell, Mielke Nakagami, non-central chi-squared, non-central F distribution, non-central Student's t, normal, Pareto, power log-normal, power normal, R-distributed, Rice, right-skewed Gumbel, semicircular, Student's t, trapezoidal, triangular, truncated exponential, truncated normal, uniform, Wald, Weibull maximum, Weibull minimum, and wrapped Cauchy continuous random variables.

Most of SRGMs were originally devised for standalone embedded systems developed using a waterfall or spiral method. They thus assume a fixed set of software faults exists from the beginning of the analysis but not yet detected. The total count of defects decreases as they are detected and fixed over time (e.g., in testing). However, with the agile method, new defects are continuously introduced, while some existing defects are automatically removed behind the scene (e.g., due to feature deprecations or changes in the user journeys) as the software evolves. It suggests us that there are more than one parameters that influence the defect counts and more complex models than the one-parameter SRGMs are necessary in order to accurately estimate the production outage counts. That is especially true because outages are due to the activations of defects as well as some operational issues.

**Implication.** *The existing SRGMs using one parameter are too simple for agile software reliability modeling. Intuitively, agile software reliability is contributed by at least two types of factors: the arrival rate of software bugs and detection/fix rates of the bugs. Those two types of factors are too distinct that they would easily diverge as the monitored time becomes longer and longer. That is, they can only be accurately modeled using more than one parameters.*

## V. DESIGN

This section describes the four types of time series forecasting (TSF) models used in this study and presents the model optimization techniques. Here, outage count forecasting is modeled as a univariate TSF problem.

**Rare, spiky production outages are difficult to predict.** Instead of modeling a software engineering process and estimating the defect counts, in this paper, we use TSF models to directly forecast the production outage counts. However, forecasting rare, spiky, and imbalanced events (i.e., a type of extreme events [18][19][20]) is still an active area of research. The types of extreme events considered in the past include: holiday Uber usage surges [21], peak wind speed [22], Nasdaq individual stock prices, and greenhouse gas and CO2 concentrations [23]. Those previously studied extreme events are, however, neither sparse nor rare compared with our target production outage events that are usually 0 if we break them down by the root cause type (or using the top-level software component).

### A. Forecasting Models

We use the four kinds of forecasting models. Here, we use a *lag* parameter to indicate the number of previous samples to lookback for prediction and a *horizon* parameter to specify the number of samples to predict.

**Previous Value (PV).** PV predicts the next value as the previous value, i.e., $o_i = o_{i-1}$.

**Moving Average (MA).** MA takes an average of the recent values and predicts the average as the next value. That is, $o_i = (\sum_{j=i-lag}^{i-1} o_j)/lag$. The PV model is a special case of the MA model where the lag is 1, i.e., PV = MA(1).

**Auto-Regressive (AR).** AR is a logistic regression model trained on recent data. AR model is formulated as: $o_i = b + (\sum_{j=1}^{lag} c_j o_{i-j}) + \varepsilon_i$. Here, $b$ is a constant variable; $\{c_i\}_{i=1}^{lag}$ is the autoregressive parameters; and $\{\varepsilon_1, \varepsilon_2, \ldots\}$ is a white noise with the zero mean value. We use the Python implementation of AR-X model[3]. The AR-X model accepts exogenous input (e.g., covariance data). The model parameters are derived by using the conditional maximum likelihood method (CML) and the ordinary least squares (OLS) method [24][25][26].

**Foundational Model (FM).** FM is a foundational model pre-trained to forecast time series data. We use the state-of-the-art TimesFM [8] as an FM in this study. TimesFM is chosen as a representative model because it is reported to perform better than the simpler models, such as LSTM (Long Short-Term Memory)- [21] and GRU (Gated Recurrent Unit)-based [23] models.

TimesFM employs a decoder-only Transformer model that well adapts to different context lengths (or lags). It employs patching to breakdown training data into patches (analogous to tokens in large language models, LLMs) for the accuracy and inference speed. For the horizon parameter of 1, the input, transformer, and output layers of TimesFM are modeled as:

$$t_i = InputResidualBlock(y_i \odot (1 - m_i)) + PE_i$$
$$o_i = StackedTransformer((t_1, m_1), \ldots, (t_i, m_i))$$
$$\hat{y}_i = OutputResidualBlock(o_i)$$

It is trained to minimize the mean squared error of $\hat{y}_i$ and $y_i$. We use the pre-trained model, HuggingFace google/timesfm-1.0-200m. The pre-trained model [8] is trained on large-scale query trends, Wikimedia hourly page views, and synthetic time series. The pre-trained mode with 200M parameters is configured with the 20 layers, 1280 dimensions, input patch length of 32, and output patch length of 128 (i.e., the same as the configuration used in [8]).

### B. Optimization

We divide a given time series into the three sub-intervals: train, validation, and test. If a model type supports training, the train interval is used to train the model and the validation interval is used to validate the trained model. AR and FM support training. The data from the test interval is then used to evaluate the model.

We evaluate the following model optimization techniques:

**Logarithmic Transformation.** This technique applies the log1p conversion. A raw sample value of $o_i$ is converted to $log(1 + o_i)$ before being used as a model input. Similarly, a raw model output value of $\bar{o}_i$ is converted to $exp(\bar{o}_i) - 1$ before being used as the output.

**0 Floor.** If a predicted value is less than 0, this technique changes the predicted value to 0 because less than 0 is not a valid output value for our target variable. It is used for AR and FM.

**Covariance.** When a model accepts covariance parameters, we use two kinds of covariance data. One is the month of year. The other is the code freeze month information (i.e., usually December). We evaluate this technique using FM.

---
[3] statsmodels.tsa.ar_model.AutoReg module.

**Fine-tuning.** We fine tune the pre-trained FM using part of our dataset as training and validation sub-datasets. Here, as the training and validation sub-datasets, we use either the time series for all root cause types or the times series for only one relevant root cause type for fine-tuning. We evaluate the resulting forecasting accuracy of both scenarios. We also fine-tune the frequency parameter of FM: 0 or 1.

## VI. RESULT

This section analyzes the evaluation results. We use either the direct single-step or iterated multi-step (IMS) forecasting because our targets are the short-term, time series forecasting scenarios. That is, the horizon parameter value is 1 month. At the right beginning of each month, using the actual values of the last month, we dynamically retrain a model to accurately predict the values for the present month to minimize the error accumulation effect and temporal information loss that vanilla Transform-based models used for long-term forecasting would exhibit [27].

### A. Predicting Monthly Total Outages

We predict the monthly total outage counts and measure the prediction errors. Table IV summarizes the prediction errors as a function of the model type and lag value. FM consistently achieves the lowest errors, with a lag of 7 months providing the best accuracy. The MA model shows the second lowest errors, with a lag of 6 months providing the second best accuracy in terms of $\overline{MAE}$. FM(7) achieves 1.3% smaller $\overline{MAE}$, 12.4% smaller $\overline{MSE}$, and 6.4% smaller $\overline{RMSE}$ than the second best model, MA(6). Here, FM employs the logarithmic transformation and 0 flooring techniques that generally lead to the higher accuracy. Despite considering various lag values and retraining monthly, the AR model is less effective for the used dataset than MA. Increasing the lag value for AR with the cap of 12 months as more samples are collected does not improve the prediction accuracy.

Figure 3(a), 3(b), and 3(c) show the predictions of FM(7), AR(2), and MA(6), respectively. The dash dotted line is for the actual outage counts (normalized). The solid line is for the forecasted outage counts (normalized). The dotted line at the bottom of each graph is for $\overline{MAE}$. The LLM-based FM(7) accurately forecasts the total outages by quickly adapting to the spikes and drops. While AR(2) quickly catches spikes and drops, it usually incorrectly predicts that the spikes will grow further by when the spikes are actually gone, resulting in the larger prediction errors than FM(7). MA(6) tracks the recent averages and thus does not fully catch recent spikes and drops if they only last for 1 or 2 months.

Using the same models to forecast the last two years of data, the MA model shows the highest accuracy (see Table V). While FM performs well when there are large variations in the 2nd year and 3rd year, MA performs well when there are small variations in the 6th and 7th years. In summary, using an FM gives the 1.3-12.4% gain on average in terms of forecasting accuracy compared with the best classical model evaluated for forecasting the total monthly outages of our entire dataset. The

TABLE IV
NORMALIZED PREDICTION ERRORS FOR THE MONTHLY TOTAL OUTAGES
(TEST DATA PERIOD: 6 YEARS FROM 2ND TO 7TH YEAR) – UNIT: 0.001

| Model Type | Lag | MAE | MSE | RMSE |
|---|---|---|---|---|
| Previous Value | 1 | 2.58735 | 0.01056 | 3.24984 |
| Moving Average | 5 | 2.12163 | 0.00900 | 3.00002 |
|  | 6 | 2.12076 | 0.00866 | 2.94222 |
|  | 7 | 2.20738 | 0.00883 | 2.97141 |
| Auto Regressive | 1 | 2.63907 | 0.01087 | 3.29660 |
|  | 2 | 2.61802 | 0.01103 | 3.32076 |
|  | 3 | 2.62402 | 0.01111 | 3.33346 |
| Foundational Model | 6 | 2.12350 | 0.00786 | 2.80433 |
|  | 7 | **2.09248** | **0.00759** | **2.75457** |
|  | 8 | 2.12604 | 0.00781 | 2.79376 |

TABLE V
NORMALIZED PREDICTION ERRORS FOR THE MONTHLY TOTAL OUTAGES
(TEST DATA PERIOD: 2 YEARS 6TH AND 7TH YEARS) – UNIT: 0.001

| Model Type | Lag | MAE | MSE | RMSE |
|---|---|---|---|---|
| Previous Value | 1 | 2.04918 | 0.00606 | 2.46203 |
| Moving Average | 7 | 1.33951 | 0.00277 | 1.66558 |
|  | 8 | 1.38358 | 0.00294 | 1.71363 |
|  | 9 | 1.32645 | 0.00283 | 1.68228 |
| Auto Regressive | 3 | 1.83171 | 0.00531 | 2.30531 |
|  | 4 | 1.76198 | 0.00489 | 2.21128 |
|  | 5 | 1.78734 | 0.00512 | 2.26243 |
| Foundational Model | 6 | 1.52462 | 0.00325 | 1.80385 |
|  | 7 | 1.53397 | 0.00339 | 1.83997 |
|  | 8 | 1.53333 | 0.00333 | 1.82534 |

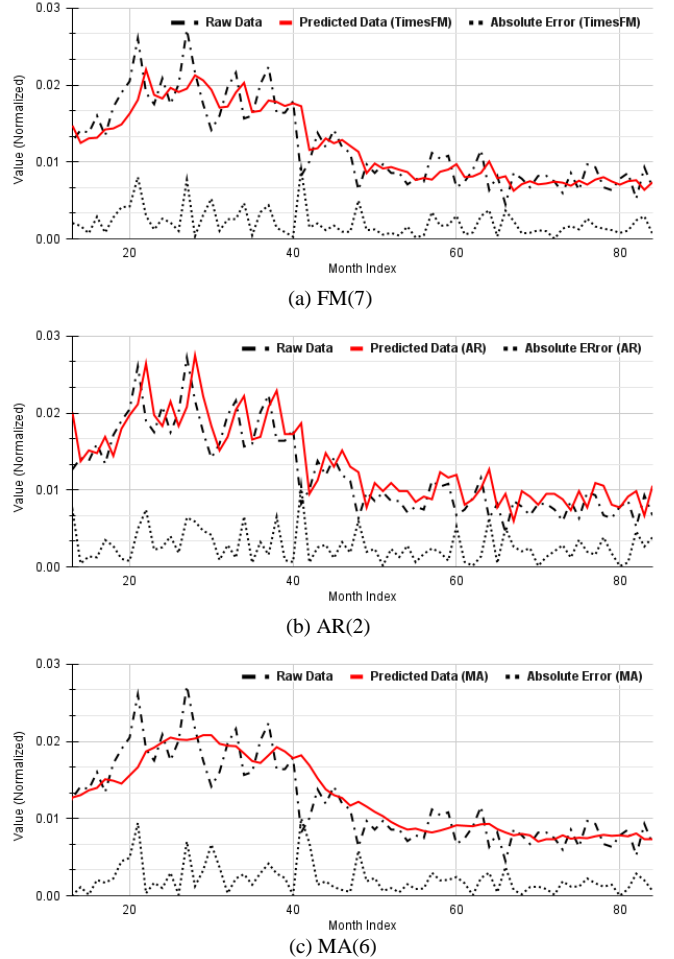

(a) FM(7)

(b) AR(2)

(c) MA(6)

**Fig. 3.** Prediction results of FM, the AR model, and the MA model.

TABLE VI
PREDICTION ERRORS FOR THE ABSOLUTE OUTAGE COUNTS OF EACH ROOT CAUSE TYPE (LAST 5 YEARS) – UNIT: 0.0001

| Model | Lag | Capacity | | Experiment | | Database | | Frontend | | Data | | ML | | Client | | Migration | | AVG | | |
|---|---|---|---|---|---|---|---|---|---|---|---|---|---|---|---|---|---|---|---|---|
| | | MAE | RMSE | MAE | RMSE | MAE | RMSE | MAE | RMSE | MAE | RMSE | MAE | RMSE | MAE | RMSE | MAE | RMSE | MAE | RMSE | Both |
| PV | 1 | 4.347 | 5.812 | 6.955 | 9.376 | 3.105 | 4.858 | 0.994 | 2.453 | 4.471 | 7.326 | 3.477 | 4.999 | 1.739 | 3.191 | 2.484 | 5.356 | 3.447 | 5.421 | 4.434 |
| MA | 2 | 3.974 | 5.388 | 5.431 | 9.422 | 2.670 | 4.152 | 1.118 | 2.381 | 3.943 | 6.413 | 3.415 | 4.408 | 1.615 | 3.004 | 2.670 | 5.732 | 3.105 | 5.113 | 4.109 |
| | 3 | 3.891 | 5.380 | 6.251 | 8.652 | 2.442 | 4.018 | 1.076 | 2.334 | 3.809 | 6.139 | 3.333 | 4.215 | 1.573 | 2.777 | 2.732 | 5.508 | 3.138 | 4.878 | 4.008 |
| | 4 | 3.974 | 5.450 | 6.505 | 8.583 | 2.484 | 4.017 | 1.056 | 2.294 | 3.695 | 6.072 | 3.245 | 4.008 | 1.490 | 2.511 | 2.639 | 5.175 | 3.136 | 4.764 | 3.950 |
| | 5 | 4.123 | 5.496 | 6.793 | 8.744 | 2.683 | 4.133 | 1.093 | 2.285 | 3.602 | 5.949 | 3.192 | 3.949 | 1.391 | 2.368 | 2.658 | 4.991 | 3.192 | 4.739 | 3.966 |
| | 6 | 4.264 | 5.630 | 6.758 | 8.883 | 2.774 | 4.080 | 1.056 | 2.222 | 3.664 | 6.010 | 3.208 | 3.880 | 1.428 | 2.460 | 2.753 | 5.198 | 3.238 | 4.795 | 4.017 |
| | 7 | 4.294 | 5.629 | 6.698 | 8.931 | 2.910 | 4.146 | 1.047 | 2.224 | 3.690 | 6.097 | 3.167 | 3.830 | 1.508 | 2.588 | 2.981 | 5.456 | 3.287 | 4.863 | 4.075 |
| | 8 | 4.300 | 5.544 | 6.229 | 7.552 | 2.950 | 4.155 | 1.087 | 2.288 | 3.726 | 6.084 | 3.260 | 3.891 | 1.490 | 2.575 | 3.151 | 5.645 | 3.274 | 4.717 | 3.995 |
| AR | 1 | 4.439 | 5.554 | 7.392 | 11.039 | 2.854 | 4.050 | 1.154 | 2.402 | 4.297 | 7.247 | 3.268 | 4.032 | 1.481 | 2.804 | 2.699 | 5.423 | 3.448 | 5.319 | 4.384 |
| | 2 | 4.442 | 5.649 | 8.801 | 15.641 | 2.778 | 4.001 | 1.748 | 4.670 | 4.072 | 6.958 | 3.369 | 4.147 | 1.474 | 2.818 | 3.010 | 6.443 | 3.712 | 6.291 | 5.001 |
| | 3 | 4.579 | 5.797 | 11.606 | 27.002 | 2.873 | 4.159 | 2.527 | 8.358 | 4.625 | 7.436 | 3.418 | 4.227 | 1.519 | 2.852 | 3.246 | 6.967 | 4.299 | 8.350 | 6.324 |
| | 4 | 4.640 | 6.059 | 18.245 | 54.872 | 2.985 | 4.444 | 3.705 | 12.958 | 4.820 | 7.662 | 3.596 | 4.385 | 1.553 | 2.794 | 3.683 | 7.538 | 5.403 | 12.589 | 8.996 |
| FM | 3 | 3.347 | 4.966 | 5.382 | 8.038 | 2.114 | 3.656 | 0.854 | 2.105 | 3.099 | 5.681 | 2.869 | 3.963 | 1.216 | 2.541 | 2.032 | 4.765 | 2.614 | 4.464 | 3.539 |
| | 4 | 3.411 | 5.075 | 5.618 | 7.981 | 2.176 | 3.709 | 0.813 | 2.077 | 3.151 | 5.754 | 2.933 | 4.048 | 1.139 | 2.479 | 2.008 | 4.643 | 2.656 | 4.471 | 3.563 |
| | 5 | 3.502 | 5.003 | 5.682 | 8.063 | 2.121 | 3.662 | 0.827 | 2.096 | 2.965 | 5.697 | 2.859 | 3.932 | 1.061 | 2.353 | 1.791 | 4.321 | 2.601 | 4.391 | 3.496 |
| | 6 | 3.576 | 5.137 | 5.667 | 8.208 | 2.109 | 3.629 | 0.833 | 2.134 | 3.001 | 5.771 | 2.790 | 3.796 | 1.186 | 2.493 | 1.839 | 4.585 | 2.625 | 4.469 | 3.547 |
| | 7 | 3.459 | 4.952 | 5.650 | 8.090 | 2.169 | 3.668 | 0.799 | 2.096 | 2.919 | 5.689 | 2.674 | 3.724 | 1.253 | 2.591 | 1.975 | 4.714 | 2.612 | 4.441 | 3.526 |
| | 8 | 3.406 | 4.941 | 5.798 | 8.187 | 2.120 | 3.631 | 0.861 | 2.177 | 2.950 | 5.702 | 2.811 | 3.858 | 1.189 | 2.503 | 1.977 | 4.738 | 2.639 | 4.467 | 3.553 |
| | 9 | 3.386 | 5.006 | 5.629 | 7.976 | 2.145 | 3.653 | 0.892 | 2.181 | 3.004 | 5.789 | 2.874 | 3.875 | 1.134 | 2.452 | 1.934 | 4.708 | 2.625 | 4.455 | 3.540 |
| | 10 | 3.358 | 4.978 | 5.712 | 8.039 | 2.151 | 3.663 | 0.909 | 2.237 | 2.956 | 5.810 | 2.936 | 3.941 | 1.173 | 2.535 | 1.985 | 4.779 | 2.648 | 4.498 | 3.573 |
| | 11 | 3.345 | 5.006 | 5.595 | 7.955 | 2.238 | 3.717 | 0.888 | 2.197 | 2.916 | 5.786 | 2.875 | 3.835 | 1.177 | 2.542 | 1.971 | 4.810 | 2.626 | 4.481 | 3.553 |
| | 12 | 3.374 | 5.123 | 5.427 | 7.756 | 2.239 | 3.728 | 0.870 | 2.195 | 2.884 | 5.779 | 2.734 | 3.711 | 1.227 | 2.598 | 1.993 | 4.877 | 2.594 | 4.471 | 3.532 |

TABLE VII
PREDICTION ERRORS FOR FORECASTING THE ABSOLUTE PART VALUE OF EACH ROOT CAUSE TYPE (LAST 2 YEARS) – UNIT: 0.0001

| Model | Lag | Capacity | | Experiment | | Database | | Frontend | | Data | | ML | | Client | | Migration | | AVG | | |
|---|---|---|---|---|---|---|---|---|---|---|---|---|---|---|---|---|---|---|---|---|
| | | MAE | RMSE | MAE | RMSE | MAE | RMSE | MAE | RMSE | MAE | RMSE | MAE | RMSE | MAE | RMSE | MAE | RMSE | MAE | RMSE | Both |
| MA | 2 | 2.484 | 4.024 | 7.219 | 9.087 | 1.397 | 2.282 | 0.000 | 0.000 | 5.589 | 8.156 | 2.484 | 3.443 | 1.941 | 3.018 | 3.493 | 5.804 | 3.076 | 4.477 | 3.776 |
| | 3 | 2.329 | 3.558 | 6.210 | 8.160 | 1.501 | 2.522 | 0.000 | 0.000 | 5.692 | 8.073 | 2.536 | 3.296 | 1.863 | 2.635 | 3.778 | 6.287 | 2.989 | 4.316 | 3.652 |
| | 4 | 2.173 | 3.348 | 6.559 | 8.277 | 1.358 | 2.176 | 0.000 | 0.000 | 5.472 | 7.983 | 2.406 | 3.193 | 1.785 | 2.344 | 3.571 | 5.589 | 2.916 | 4.114 | 3.515 |
| | 5 | 2.422 | 3.381 | 6.862 | 8.271 | 1.615 | 2.386 | 0.031 | 0.152 | 5.433 | 7.832 | 2.329 | 3.046 | 1.708 | 2.261 | 3.571 | 5.142 | 2.996 | 4.059 | 3.527 |
| | 6 | 2.536 | 3.607 | 6.675 | 8.134 | 1.578 | 2.221 | 0.052 | 0.179 | 5.511 | 7.878 | 2.380 | 2.994 | 1.811 | 2.425 | 3.855 | 5.616 | 3.050 | 4.132 | 3.591 |
| | 7 | 2.617 | 3.524 | 6.387 | 7.851 | 1.730 | 2.360 | 0.067 | 0.188 | 5.633 | 8.085 | 2.284 | 2.827 | 1.952 | 2.594 | 4.347 | 6.148 | 3.127 | 4.197 | 3.662 |
| | 8 | 2.736 | 3.584 | 6.210 | 7.475 | 1.708 | 2.274 | 0.097 | 0.252 | 5.822 | 8.077 | 2.561 | 3.075 | 1.979 | 2.576 | 4.715 | 6.555 | 3.229 | 4.234 | 3.731 |
| AR | 1 | 3.309 | 4.110 | 5.837 | 8.193 | 2.252 | 2.551 | 0.404 | 0.407 | 6.241 | 9.172 | 2.810 | 3.133 | 2.080 | 3.038 | 3.133 | 5.046 | 3.258 | 4.456 | 3.857 |
| | 2 | 3.086 | 3.909 | 5.894 | 8.170 | 1.859 | 2.245 | 0.381 | 0.384 | 5.739 | 8.630 | 2.761 | 3.096 | 2.081 | 3.061 | 3.163 | 5.108 | 3.121 | 4.325 | 3.723 |
| | 3 | 3.072 | 3.902 | 5.378 | 7.556 | 1.930 | 2.372 | 0.343 | 0.346 | 6.442 | 9.055 | 2.806 | 3.129 | 1.976 | 2.957 | 4.261 | 8.416 | 3.276 | 4.717 | 3.996 |
| | 4 | 3.105 | 3.952 | 5.389 | 7.607 | 2.003 | 2.466 | 0.312 | 0.314 | 6.611 | 9.156 | 2.759 | 3.178 | 2.091 | 2.713 | 4.826 | 8.786 | 3.387 | 4.772 | 4.079 |
| FM (Freq=0) | 5 | 2.327 | 3.588 | 6.592 | 8.150 | 1.501 | 2.309 | 0.001 | 0.002 | 5.358 | 8.043 | 2.490 | 3.400 | 1.531 | 2.605 | 2.026 | 4.070 | 2.728 | 4.021 | 3.375 |
| | 6 | 2.330 | 3.640 | 6.517 | 8.061 | 1.502 | 2.242 | 0.003 | 0.011 | 5.395 | 8.131 | 2.415 | 3.270 | 1.791 | 2.825 | 2.292 | 4.929 | 2.781 | 4.139 | 3.460 |
| | 7 | 2.269 | 3.447 | 6.446 | 8.068 | 1.587 | 2.294 | 0.009 | 0.010 | 5.230 | 7.969 | 2.207 | 3.051 | 1.952 | 2.953 | 2.696 | 5.225 | 2.800 | 4.127 | 3.463 |
| | 8 | 2.326 | 3.450 | 6.538 | 8.101 | 1.452 | 2.116 | 0.002 | 0.010 | 5.312 | 7.963 | 2.395 | 3.252 | 1.761 | 2.699 | 2.690 | 5.317 | 2.810 | 4.114 | 3.462 |
| | 9 | 2.387 | 3.465 | 6.374 | 7.843 | 1.496 | 2.140 | 0.006 | 0.009 | 5.351 | 8.097 | 2.319 | 3.123 | 1.648 | 2.591 | 2.562 | 5.277 | 2.768 | 4.068 | 3.418 |
| | 10 | 2.368 | 3.420 | 6.285 | 7.704 | 1.452 | 2.080 | 0.003 | 0.007 | 5.264 | 8.141 | 2.363 | 3.138 | 1.721 | 2.718 | 2.740 | 5.430 | 2.775 | 4.080 | 3.427 |
| | 11 | 2.338 | 3.396 | 6.113 | 7.536 | 1.569 | 2.210 | 0.005 | 0.012 | 5.090 | 8.043 | 2.342 | 3.101 | 1.729 | 2.710 | 2.729 | 5.498 | 2.739 | 4.063 | 3.401 |

optimal model type and lag value depend on the characteristics of target data if only part of our dataset is used as a forecasting target.

### B. Predicting Monthly Outages of Each Root Cause Type

We predict the monthly outage counts of each root cause type. Table VI summarizes the prediction accuracy errors for each root cause type as a function of the prediction model and lag parameter value. FM shows the highest accuracy for 7 root cause types. The exception is the experiment root cause type where MA(8) shows the lower $\overline{RMSE}$ than FM but the higher $\overline{MAE}$ than FM. On average, FM shows 19.7% higher accuracy than the second best model when $\overline{MAE}$ is used as an error metric and 7.4% higher accuracy when $\overline{RMSE}$ is an error metric.

When using the last two years of data as the test dataset, the result is different. As summarized in Table VII, MA shows the higher accuracy than AR and FM for the capacity, database, frontend, and ML root cause types. FM shows the highest accuracy for the migration root cause type. AR shows the highest accuracy for the experiment root cause type. MA also shows the smaller $\overline{RMSE}$ than the two other models for the data and client root cause types, while FM shows the smaller $\overline{MAE}$ than the other two model types for the same two root cause types.

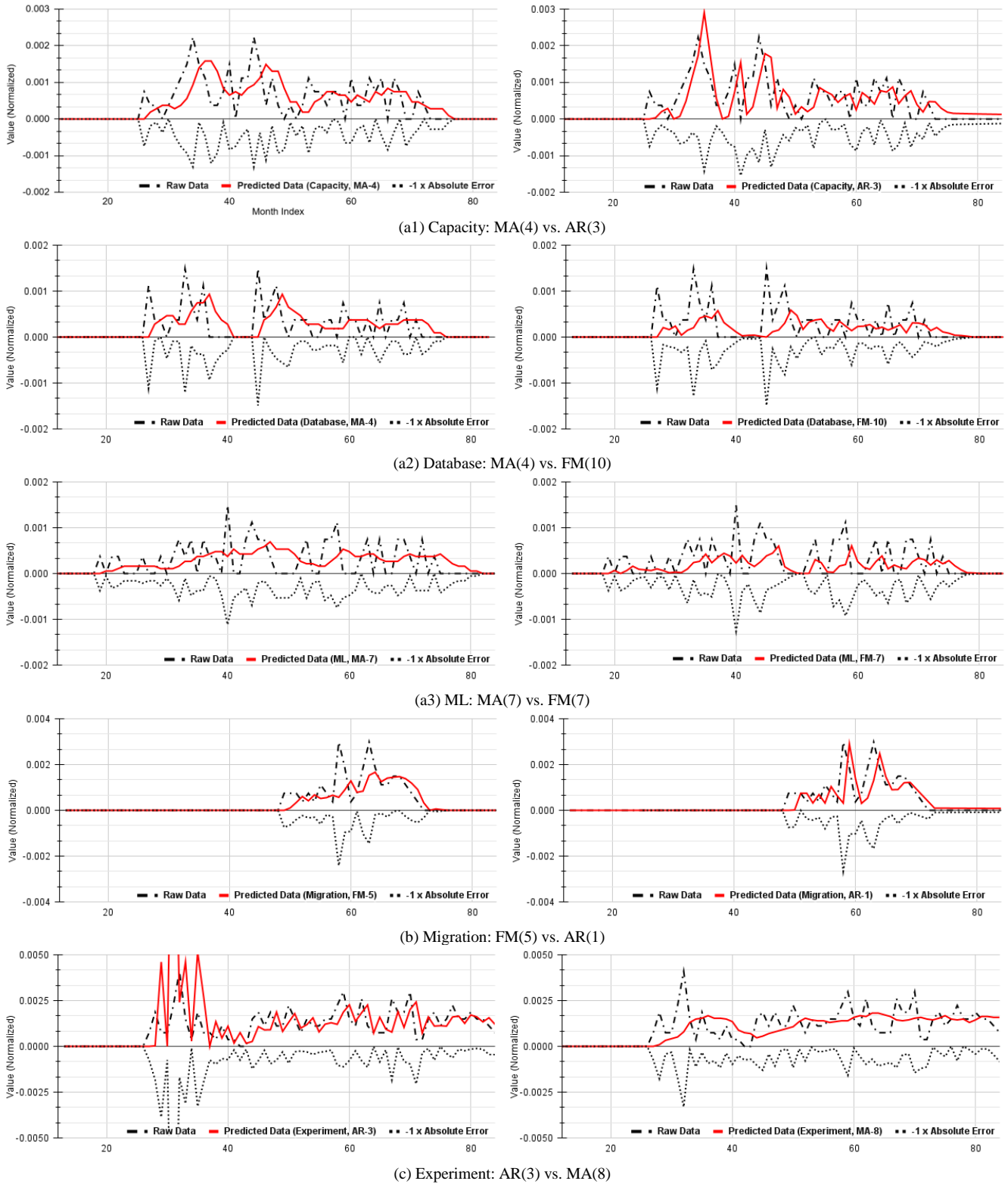

(a1) Capacity: MA(4) vs. AR(3)

(a2) Database: MA(4) vs. FM(10)

(a3) ML: MA(7) vs. FM(7)

(b) Migration: FM(5) vs. AR(1)

(c) Experiment: AR(3) vs. MA(8)

**Fig. 4.** Prediction results for each root cause type (x-axis: week index).

Figure 4 visualizes them. For capacity type, AR does not catch that the reference value becomes 0. For database type, MA captures the median well since the data in the 6$^{th}$ year is somewhat stationary, while FM does not capture the moving averages well. Also, MA converges to 0 much quicker than FM in the 7$^{th}$ year. For ML data type, FM converged to 0 quicker than MA(7) that uses the longer lag value than MA(4) used for database data type. The MA(7) model well tracks the average of peaks. Since the last peak is similar to the average of the previous peaks, MA(7) more accurately estimates the target values than FM that does not track the spikes well.

TABLE VIII
PREDICTION ERRORS FOR FORECASTING THE ABSOLUTE PART VALUE OF EACH ROOT CAUSE TYPE USING FINE-TUNING (LAST 2 YEARS) – UNIT: 0.0001

| Model | Lag | Capacity | | Experiment | | Database | | Frontend | | Data | | ML | | Client | | Migration | | AVG |
|---|---|---|---|---|---|---|---|---|---|---|---|---|---|---|---|---|---|---|
| | | nMAE | nRMSE | nMAE | nRMSE | nMAE | nRMSE | nMAE | nRMSE | nMAE | nRMSE | nMAE | nRMSE | nMAE | nRMSE | nMAE | nRMSE | |
| TimesFM (Freq=0) | 5 | 2.327 | 3.588 | 6.592 | 8.150 | 1.501 | 2.309 | 0.001 | 0.002 | 5.358 | 8.043 | 2.490 | 3.400 | 1.531 | 2.605 | 2.026 | 4.070 | 3.375 |
| | 6 | 2.330 | 3.640 | 6.517 | 8.061 | 1.502 | 2.242 | 0.003 | 0.011 | 5.395 | 8.131 | 2.415 | 3.270 | 1.791 | 2.825 | 2.292 | 4.929 | 3.460 |
| | 7 | 2.269 | 3.447 | 6.446 | 8.068 | 1.587 | 2.294 | 0.009 | 0.010 | 5.230 | 7.969 | 2.207 | 3.051 | 1.952 | 2.953 | 2.696 | 5.225 | 3.463 |
| | 8 | 2.326 | 3.450 | 6.538 | 8.101 | 1.452 | 2.116 | 0.002 | 0.010 | 5.312 | 7.963 | 2.395 | 3.252 | 1.761 | 2.699 | 2.690 | 5.317 | 3.462 |
| | 9 | 2.387 | 3.465 | 6.374 | 7.843 | 1.496 | 2.140 | 0.006 | 0.009 | 5.351 | 8.097 | 2.319 | 3.123 | 1.648 | 2.591 | 2.562 | 5.277 | 3.418 |
| | 10 | 2.368 | 3.420 | 6.285 | 7.704 | 1.452 | 2.080 | 0.003 | 0.007 | 5.264 | 8.141 | 2.363 | 3.138 | 1.721 | 2.718 | 2.740 | 5.430 | 3.427 |
| | 11 | 2.338 | 3.396 | 6.113 | 7.536 | 1.569 | 2.210 | 0.005 | 0.012 | 5.090 | 8.043 | 2.342 | 3.101 | 1.729 | 2.710 | 2.729 | 5.498 | 3.401 |
| TimesFM (Fine-Tune, Freq=0) | 5 | 2.329 | 3.589 | 6.597 | 8.155 | 1.500 | 2.307 | 0.000 | 0.002 | 5.359 | 8.043 | 2.488 | 3.397 | **1.529** | 2.603 | **2.017** | 4.069 | **3.374** |
| | 6 | 2.328 | 3.640 | 6.520 | 8.065 | 1.502 | 2.242 | 0.002 | 0.010 | 5.399 | 8.134 | 2.413 | 3.268 | 1.787 | 2.822 | 2.289 | 4.926 | 3.459 |
| | 7 | 2.264 | 3.443 | 6.446 | 8.066 | 1.585 | 2.294 | 0.001 | 0.004 | 5.231 | 7.972 | **2.204** | 3.051 | 1.949 | 2.954 | 2.692 | 5.222 | 3.461 |
| | 8 | 2.322 | 3.447 | 6.541 | 8.100 | 1.453 | 2.118 | 0.002 | 0.009 | 5.310 | 7.964 | 2.393 | 3.250 | 1.757 | 2.697 | 2.689 | 5.316 | 3.461 |
| | 9 | 2.385 | 3.463 | 6.377 | 7.844 | 1.496 | 2.140 | 0.002 | 0.005 | 5.351 | 8.099 | 2.317 | 3.121 | 1.645 | **2.589** | 2.556 | 5.275 | 3.417 |
| | 10 | 2.366 | 3.417 | 6.289 | 7.706 | 1.451 | **2.080** | 0.002 | 0.005 | 5.264 | 8.143 | 2.359 | 3.134 | 1.719 | 2.717 | 2.734 | 5.427 | 3.426 |
| | 11 | 2.335 | 3.392 | 6.117 | 7.539 | 1.570 | 2.211 | 0.004 | 0.011 | **5.090** | 8.046 | 2.340 | 3.098 | 1.726 | 2.707 | 2.729 | 5.498 | 3.401 |
| TimesFM (Fine-Tune, Freq=1) | 5 | 2.268 | 3.615 | 6.343 | 8.020 | 1.581 | 2.437 | **0.000** | **0.000** | 5.278 | 7.906 | 2.516 | 3.467 | 1.619 | 2.686 | 2.377 | 4.383 | 3.406 |
| | 6 | 2.393 | 3.762 | 6.066 | 7.818 | 1.536 | 2.320 | 0.001 | 0.004 | 5.262 | 7.980 | 2.439 | 3.373 | 1.811 | 2.853 | 2.728 | 5.241 | 3.474 |
| | 7 | 2.298 | 3.549 | 6.015 | 7.745 | 1.612 | 2.374 | 0.000 | 0.001 | 5.138 | 7.940 | 2.269 | 3.196 | 1.944 | 2.920 | 2.691 | 5.202 | 3.431 |
| | 8 | 2.293 | 3.489 | 6.133 | 7.879 | 1.469 | 2.165 | 0.001 | 0.002 | 5.146 | **7.836** | 2.539 | 3.502 | 1.840 | 2.771 | 2.615 | 5.134 | 3.426 |
| | 9 | 2.330 | 3.470 | 6.029 | 7.742 | 1.507 | 2.167 | 0.001 | 0.004 | 5.267 | 7.978 | 2.394 | 3.289 | 1.787 | 2.709 | 2.590 | 5.210 | 3.405 |
| | 10 | 2.266 | 3.353 | 5.966 | 7.660 | 1.488 | 2.153 | 0.003 | 0.008 | 5.230 | 8.111 | 2.361 | 3.210 | 1.872 | 2.852 | 2.704 | 5.278 | 3.407 |
| | 11 | **2.232** | **3.351** | **5.822** | **7.531** | 1.573 | 2.250 | 0.004 | 0.010 | 5.145 | 8.008 | 2.324 | 3.154 | 1.809 | 2.858 | 2.787 | 5.354 | 3.388 |
| TimesFM (Fine-Tune per data, Freq=0) | 5 | 2.700 | 4.005 | 6.798 | 8.469 | 1.615 | 2.448 | 0.001 | 0.003 | 5.358 | 8.046 | 2.495 | 3.387 | 1.534 | 2.606 | 2.062 | **4.068** | 3.475 |
| | 6 | 2.711 | 3.904 | 7.242 | 8.831 | 1.534 | 2.261 | 0.003 | 0.011 | 5.409 | 8.145 | 2.427 | 3.266 | 1.792 | 2.824 | 2.311 | 4.877 | 3.597 |
| | 7 | 2.529 | 3.391 | 6.853 | 8.313 | **1.410** | 2.099 | 0.009 | 0.009 | 5.232 | 7.982 | 2.217 | **3.043** | 1.954 | 2.955 | 2.717 | 5.194 | 3.494 |
| | 8 | 2.474 | 3.388 | 7.104 | 8.401 | 1.472 | 2.098 | 0.002 | 0.010 | 5.309 | 7.975 | 2.403 | 3.247 | 1.764 | 2.704 | 2.689 | 5.289 | 3.521 |
| | 9 | 2.505 | 3.345 | 7.057 | 8.209 | 1.559 | 2.200 | 0.007 | 0.009 | 5.359 | 8.108 | 2.325 | 3.111 | 1.649 | 2.591 | 2.558 | 5.258 | 3.491 |
| | 10 | 2.490 | 3.311 | 6.707 | 7.726 | 1.549 | 2.160 | 0.003 | 0.007 | 5.270 | 8.152 | 2.371 | 3.131 | 1.719 | 2.708 | 2.767 | 5.435 | 3.469 |
| | 11 | 2.444 | 3.283 | 6.329 | 7.280 | 1.498 | 2.187 | 0.005 | 0.012 | 5.090 | 8.054 | 2.349 | 3.091 | 1.728 | 2.707 | 2.718 | 5.473 | 3.391 |

Figure 4(b) shows when FM performs the best. The migration-oriented outage counts in the 6[th] year have a relatively smoother spike (i.e., slightly longer duration than shape spikes) that gradually decreases to 0. That pattern is well tracked by FM. AR is the second best as it loses some errors because it does not completely converge to 0 and more spontaneously reacting to the drop and spike at the beginning of the 6[th] year than FM.

Finally, Figure 4(c) shows when AR shows the least errors. The experiment outage counts have double spikes that are relatively well tracked by AR because AR tracks the first spike with some time delay by when the second spike actually rises. The target data is then followed by a trending up line with three small spikes that are all well captured by AR. This analysis helps us understand what kinds of time series patterns are well captured by each type of the used models.

**Predicting Monthly Outage Counts of Each Root Cause Type with Fine-Tuning.** When we have sufficient historical data, a natural next step to optimize the accuracy is fine-tuning a pre-trained FM. We thus select the first five years of data for fine-tuning (where the first four years of data are used for training and the fifth year data is used for validation) and the last two years of data for testing. We use the root cause type-specific time series (all eight types) but not the total monthly counts time series for fine-tuning.

Table VIII shows the results with and without fine-tuning. On average, fine-tuning gives a slight gain (0.00033745625 vs. 0.0003374) which is less than a 0.02% decrease in the average normalized errors. We then fine-tune TimesFM using the frequency of 1 (instead of 0). Changing the frequency does not provide an accuracy gain in terms of the average errors. However, in at least three root cause types, fine-tuning using frequency of 1 provides some clear accuracy gains. In that case, the optimal lag is often much longer (11 vs. 5) than that of the baseline pre-trained model and the fine-tuned model with frequency of 0.

We also fine-tune each root cause type by only using the historical data of a respective root cause type. Since the used data is small (~60 sample values for training and validation), it does not result in any accuracy gains. Thus, fine-tuning using a small, narrow sample set is not useful for our target data. The result is aligned with the previous observation [28] that a pre-trained transformer generally shows the high accuracy in the time series forecasting tasks. That is true when the model pre-trained on language and vision data is fine-turned in a restrictive way. That is, the self-attention and feed-forward layers should not be adjusted during the fine-tuning step in order to achieve the higher accuracy than a pre-trained model.

We note that using covariance (such as month of year and code freeze month) does not help with the forecasting accuracy for the used FM. That is because the used TimesFM also has the data granularity information (e.g., Month) as part of the dataset specification, and the code freeze is always in a certain year of month. Thus, they are not new information for the model.

TABLE IX
PREDICTED EXPERIMENT-CAUSED OUTAGES IN 'B'-TH MONTH OF THE FISCAL YEAR USING THE PRESENTED APPROACH USING THE AR(3), MA(8), AND FM(11) MODELS AT 'A' MONTHS BEFORE (UNIT: 0.001)

(a) Actual

| A \ B | 1 | 2 | 3 | 4 | 5 | 6 | 7 | 8 | 9 | 10 | 11 | 12 | Sum | Error% |
|---|---|---|---|---|---|---|---|---|---|---|---|---|---|---|
| 0 | 1.5 | 1.9 | 1.1 | 1.5 | 2.2 | 1.5 | 1.5 | 1.9 | 1.5 | 1.5 | 1.1 | 0.7 | 17.88 | 0% |

(b) AR(3)

| A \ B | 1 | 2 | 3 | 4 | 5 | 6 | 7 | 8 | 9 | 10 | 11 | 12 | Sum | Error% |
|---|---|---|---|---|---|---|---|---|---|---|---|---|---|---|
| -12 | *1.6* | *1.1* | *1.1* | *1.1* | *1.6* | *1.7* | *1.3* | *1.7* | *1.6* | *1.3* | *1.6* | *1.2* | 16.87 | -6% |
| -11 | 1.5 | *1.1* | *1.1* | *1.1* | *1.6* | *1.7* | *1.3* | *1.7* | *1.6* | *1.3* | *1.6* | *1.2* | 16.75 | -6% |
| -10 | 1.5 | 1.9 | *1.1* | *1.1* | *1.6* | *1.7* | *1.3* | *1.7* | *1.6* | *1.3* | *1.6* | *1.2* | 17.51 | -2% |
| -9 | 1.5 | 1.9 | 1.1 | *1.1* | *1.6* | *1.7* | *1.3* | *1.7* | *1.6* | *1.3* | *1.6* | *1.2* | 17.50 | -2% |
| -8 | 1.5 | 1.9 | 1.1 | 1.5 | *1.6* | *1.7* | *1.3* | *1.7* | *1.6* | *1.3* | *1.6* | *1.2* | 17.89 | 0% |
| -7 | 1.5 | 1.9 | 1.1 | 1.5 | 2.2 | *1.7* | *1.3* | *1.7* | *1.6* | *1.3* | *1.6* | *1.2* | 18.52 | 4% |
| -6 | 1.5 | 1.9 | 1.1 | 1.5 | 2.2 | 1.5 | *1.3* | *1.7* | *1.6* | *1.3* | *1.6* | *1.2* | 18.35 | 3% |
| -5 | 1.5 | 1.9 | 1.1 | 1.5 | 2.2 | 1.5 | 1.5 | *1.7* | *1.6* | *1.3* | *1.6* | *1.2* | 18.59 | 4% |
| -4 | 1.5 | 1.9 | 1.1 | 1.5 | 2.2 | 1.5 | 1.5 | 1.9 | *1.6* | *1.3* | *1.6* | *1.2* | 18.74 | 5% |
| -3 | 1.5 | 1.9 | 1.1 | 1.5 | 2.2 | 1.5 | 1.5 | 1.9 | 1.5 | *1.3* | *1.6* | *1.2* | 18.62 | 4% |
| -2 | 1.5 | 1.9 | 1.1 | 1.5 | 2.2 | 1.5 | 1.5 | 1.9 | 1.5 | 1.5 | *1.6* | *1.2* | 18.78 | 5% |
| -1 | 1.5 | 1.9 | 1.1 | 1.5 | 2.2 | 1.5 | 1.5 | 1.9 | 1.5 | 1.5 | 1.1 | *1.2* | 18.33 | 3% |

(c) MA(8)

| A \ B | 1 | 2 | 3 | 4 | 5 | 6 | 7 | 8 | 9 | 10 | 11 | 12 | Sum | Error% |
|---|---|---|---|---|---|---|---|---|---|---|---|---|---|---|
| -12 | 1.5 | 1.4 | 1.5 | 1.7 | 1.5 | 1.4 | 1.5 | 1.3 | 1.4 | 1.6 | 1.6 | 1.6 | 18.16 | 2% |
| -11 | 1.5 | 1.4 | 1.5 | 1.7 | 1.5 | 1.4 | 1.5 | 1.3 | 1.4 | 1.6 | 1.6 | 1.6 | 18.16 | 2% |
| -10 | 1.5 | 1.9 | 1.5 | 1.7 | 1.5 | 1.4 | 1.5 | 1.3 | 1.4 | 1.6 | 1.6 | 1.6 | 18.58 | 4% |
| -9 | 1.5 | 1.9 | 1.1 | 1.7 | 1.5 | 1.4 | 1.5 | 1.3 | 1.4 | 1.6 | 1.6 | 1.6 | 18.16 | 2% |
| -8 | 1.5 | 1.9 | 1.1 | 1.5 | 1.5 | 1.4 | 1.5 | 1.3 | 1.4 | 1.6 | 1.6 | 1.6 | 17.98 | 1% |
| -7 | 1.5 | 1.9 | 1.1 | 1.5 | 2.2 | 1.4 | 1.5 | 1.3 | 1.4 | 1.6 | 1.6 | 1.6 | 18.72 | 5% |
| -6 | 1.5 | 1.9 | 1.1 | 1.5 | 2.2 | 1.5 | 1.5 | 1.3 | 1.4 | 1.6 | 1.6 | 1.6 | 18.77 | 5% |
| -5 | 1.5 | 1.9 | 1.1 | 1.5 | 2.2 | 1.5 | 1.5 | 1.3 | 1.4 | 1.6 | 1.6 | 1.6 | 18.77 | 5% |
| -4 | 1.5 | 1.9 | 1.1 | 1.5 | 2.2 | 1.5 | 1.5 | 1.9 | 1.4 | 1.6 | 1.6 | 1.6 | 19.33 | 8% |
| -3 | 1.5 | 1.9 | 1.1 | 1.5 | 2.2 | 1.5 | 1.5 | 1.9 | 1.5 | 1.6 | 1.6 | 1.6 | 19.37 | 8% |
| -2 | 1.5 | 1.9 | 1.1 | 1.5 | 2.2 | 1.5 | 1.5 | 1.9 | 1.5 | 1.5 | 1.6 | 1.6 | 19.23 | 8% |
| -1 | 1.5 | 1.9 | 1.1 | 1.5 | 2.2 | 1.5 | 1.5 | 1.9 | 1.5 | 1.5 | 1.1 | 1.6 | 18.72 | 5% |

(d) FM(11)

| A \ B | 1 | 2 | 3 | 4 | 5 | 6 | 7 | 8 | 9 | 10 | 11 | 12 | Sum | Error% |
|---|---|---|---|---|---|---|---|---|---|---|---|---|---|---|
| -12 | 0.7 | 1.1 | 1.4 | 1.2 | 1.3 | 1.8 | 1.5 | 1.5 | 1.5 | 1.4 | 1.4 | 1.4 | 16.25 | -9% |
| -11 | 1.5 | *1.1* | 1.4 | 1.2 | 1.3 | 1.8 | 1.5 | 1.5 | 1.5 | 1.4 | 1.4 | 1.4 | 17.05 | -5% |
| -10 | 1.5 | 1.9 | 1.4 | 1.2 | 1.3 | 1.8 | 1.5 | 1.5 | 1.5 | 1.4 | 1.4 | 1.4 | 17.84 | 0% |
| -9 | 1.5 | 1.9 | 1.1 | 1.2 | 1.3 | 1.8 | 1.5 | 1.5 | 1.5 | 1.4 | 1.4 | 1.4 | 17.55 | -2% |
| -8 | 1.5 | 1.9 | 1.1 | 1.5 | 1.3 | 1.8 | 1.5 | 1.5 | 1.5 | 1.4 | 1.4 | 1.4 | 17.86 | 0% |
| -7 | 1.5 | 1.9 | 1.1 | 1.5 | 2.2 | 1.8 | 1.5 | 1.5 | 1.5 | 1.4 | 1.4 | 1.4 | 18.83 | 5% |
| -6 | 1.5 | 1.9 | 1.1 | 1.5 | 2.2 | 1.5 | 1.5 | 1.5 | 1.5 | 1.4 | 1.4 | 1.4 | 18.52 | 4% |
| -5 | 1.5 | 1.9 | 1.1 | 1.5 | 2.2 | 1.5 | 1.5 | 1.5 | 1.5 | 1.4 | 1.4 | 1.4 | 18.50 | 3% |
| -4 | 1.5 | 1.9 | 1.1 | 1.5 | 2.2 | 1.5 | 1.5 | 1.9 | 1.5 | 1.4 | 1.4 | 1.4 | 18.85 | 5% |
| -3 | 1.5 | 1.9 | 1.1 | 1.5 | 2.2 | 1.5 | 1.5 | 1.9 | 1.5 | 1.4 | 1.4 | 1.4 | 18.79 | 5% |
| -2 | 1.5 | 1.9 | 1.1 | 1.5 | 2.2 | 1.5 | 1.5 | 1.9 | 1.5 | 1.5 | 1.4 | 1.4 | 18.85 | 5% |
| -1 | 1.5 | 1.9 | 1.1 | 1.5 | 2.2 | 1.5 | 1.5 | 1.9 | 1.5 | 1.5 | 1.1 | 1.4 | 18.54 | 4% |

*C. Estimating Year-End Outage Counts*

We estimate the total outage counts until the end of the last fiscal year (i.e., the 7$^{th}$ year in our dataset) in order to assess the impact of the mitigation efforts recently put into. Table IX shows monthly estimations for the last fiscal year. We use the experiment root cause type as an example due to its relatively high outage volume in the previous year (the 6$^{th}$ year). It is a focus area of that last fiscal year. Here, iterated multi-step forecasting method is used.

AR(3) shows relatively high estimation errors (e.g., 6% = 100% − 94% in the first month, see A = -12 months in Table IX(b)). In Table IX(b), IX(c), and IX(d), the *Error*% column values are based on the actual value shown in Table IX(a). The accuracy improves slightly as more data becomes available (e.g., 0-5% error when it is less than or equal to 8 months in advance). Those low errors help us validate the effectiveness of the mitigation efforts throughout the year. Similarly, using the second and third best models for experiment type show the estimation error range of 1-8% for MA(8) and 0-9% for FM(11).

## VII. RELATED WORK

Understanding the production outages of computer network infrastructure is an important critical part of the assessment and optimization of computer system and software reliability [46][75]. Thus, forecasting models for production outages have been extensively studied.

**Individual Outage Event Prediction.** Outage-Watch [17] predicts next outage events for early detection. It monitors a set of QoS (quality of service) metrics, encodes them using bidirectional LSTM [29], and detects outages using a multi-task model trained on historical data. Other works like AirAlert [30], eWarn [31], and Fog of War [32] use alerts as features of Bayesian network- or decision tree-style classifiers. In such works, shortening the mean time to detection (MTTD) is a major challenge (i.e., long horizon prediction).

Such techniques are typically trained via supervised learning. For example, [30] and [32] use supervised learning to forecast individual events. Using autoencoder and Transformer, [34] detect faults in an industrial process captured as Tennessee Eastman benchmark. Reference [34] showed its Transformer model is better than the Deep CNN model [35]. Another technique [36] uses a transfer learning process for forecasting system metrics where the partner model is based on Random Forrest. While other existing works target system-level events, some other works [37][38][39] targets a specific hardware component, i.e., hard disk drive failures. Unlike those previous works, we forecast the short- and long-term outage trends (e.g., monthly). We also do not use any extra data other than the target data for model training.

Reference [33] is a kind of shapelet discovery technique. It forecasts software performance degradations or anomalies. In other fields of science (e.g., robotics), anomaly detection is a common challenge. Using LSTM, [40] detects anomaly in the multimodal data of a robot. Using RNN, [41] also detects anomaly in multivariate time series, while [42] uses Transformer for the same purposes.

**Time Series Forecasting.** We classify the existing time series forecasting works into the three categories:

*Statistical Methods.* ARIMA (Autoregressive Integrated Moving Average) [26] is a traditional stochastic process that consists of the AR and MA models evaluated separately in this study. It also involves transforming the target process stationary by using a logarithmic transformation of the target data and other techniques. This study characterizes whether the target data is stationary and uses the log1p transformation as an optimization technique. ARIMA is effective for short-term univariate non-stationary time series forecasting such as predicting next-day electricity price [43], wind speed, wind power generation [44], stock price [45], and cloud compute workflow [46]. Holt-Winters seasonal method [47] is another example that uses exponentially weighted moving averages.

*Machine Learning Methods.* An early work, GBERT [48], exploits gradient boosting of regression trees to produce good results. Techniques leveraging deep learning techniques such as recurrent neural network (RNN) [49], convolutional neural network (CNN), GRU [50], and LSTM [51] (e.g., LSTNet [52], TCN [53] and SCINet [54]) are studied to tackle data with a mixture of long- and short-term patterns that the traditional AR and Gaussian process may not model well.

*Transformer-based Methods.* Transformer [55] has shown its remarkable ability for the natural language processing and computer vision tasks. Its ability to capture dependencies in long historical data is a strength for some other tasks, such as time series forecasting, anomaly detection, and classification. Thus, many Transformer-based time series forecasting models have been studied. Specifically, GPT4TS [28], LLM4TS [56], and LLMTime [57] leverage models pre-trained on generic texts and vision data and show strengths (e.g., in zero-shot settings). Other specialized models trained on time series data include: PatchTST [58], LogTrans [59], Informer [60], Pyraformer [61], Triformer [62], FEDformer [63], Chronos [64], and Autoformer [65]. On the other hand, the self-attention mechanism of Transformer has the time complexity of $O(N^2)$ and thus become the computational bottleneck for long sequences [66]. LTSF-Linear (or DLinear) [27] and TSMixer [67] show the self-attention mechanism loses the temporal order and thus can show the lower accuracy for long-horizon forecasting than the linear models. The weakness is addressed by SAMformer [68]; Transformer-based time series models are an active research area and continuously evolving.

**Datasets for Transformer-based, Forecasting Models.** The existing Transformer-based models are designed for various scenarios: (1) univariate or multivariate and (2) short-term, long-term, or mixed data forecasting. However, datasets containing extreme events were not widely used to evaluate and optimize the Transformer-based forecasting models. Commonly used datasets include: electricity datasets (Electricity Transformer Temperature [69], Electricity Consumption Load [70]), stock datasets (Nasdaq Stock Market), weather datasets (temperature, humidity [71]), climate datasets (Green Gas Observing Network Dataset and Atmospheric Co2 Data), health datasets (influenza-like patents), traffic datasets, software workload datasets (app flow), and synthetic datasets. In the past, some non-Transformer models, such as Extreme Value Loss (EVL) [23], have been specifically designed for extreme events (e.g., wind speed). As far as we know, this paper is the first work that evaluates and characterizes the Transformer-based forecasting model against an extreme events dataset.

**Automation Root Cause Identification.** Our work relies on classifying outages by the root cause types. There are many existing techniques that can be used to further automate our method. For example, one may adopt an outage localization technique [72] or an outage root cause ranking technique [73] and infer the root cause types using the identified or ranked fault location information. To directly identify root cause types, one may use outage root cause identification technique [74]. Similarly, [75] can be used to select high-severity outages and generates summary texts for on-call engineers to quickly identify the root cause type(s). Using such techniques can help us use the outage count forecasting methods in near real-time.

VIII. CONCLUSION

This study analyzed the accuracy of foundational model and classical stochastic models in predicting the production outage counts of a large-scale computer software service. The main finding is that while the used foundational model on average performs the best for our target datasets, the optimal model type and lag value heavily depend on the specific patterns of the target time series data. It also showed some optimizations (e.g., logarithmic transformation and 0 flooring) are always effective, while other techniques (e.g., fine-tuning) have only marginal gains or are not effective for the used foundational model. The analysis result suggests future research directions: auto-selecting an optimal model type and auto-tuning the lag and other associated parameter values given a target dataset.

ACKNOWLEDGEMENT

The author thanks the engineering leaders who reviewed the earlier version of this paper.